\documentclass[conference]{IEEEtran}
\IEEEoverridecommandlockouts
\usepackage{cite}
\usepackage{amsmath,amssymb,amsfonts}
\usepackage{algorithm}
\usepackage{algorithmic}
\usepackage{graphicx}
\usepackage{textcomp}
\usepackage{xcolor}
\usepackage{amsmath}
\def\BibTeX{{\rm B\kern-.05em{\sc i\kern-.025em b}\kern-.08em
    T\kern-.1667em\lower.7ex\hbox{E}\kern-.125emX}}
\begin{document}

\title{VARS: Vision-based Assessment of Risk in Security Systems\\
}

\author{
\IEEEauthorblockN{Pranav Gupta, Pratham Gohil, Sridhar S}
\IEEEauthorblockA{School of Computing, SRM Institute of Science and Technology, Kattankulathur, Tamil Nadu–603203, India \\
pm4043@srmist.edu.in, ph7215@srmist.edu.in, sridhars@srmist.edu.in}
}

\maketitle

\begin{abstract}
The accurate prediction of danger levels in video content is critical for enhancing safety and security systems, particularly in environments where quick and reliable assessments are essential. In this study, we perform a comparative analysis of various machine learning and deep learning models to predict danger ratings in a custom dataset of 100 videos, each containing 50 frames, annotated with human-rated danger scores ranging from 0 to 10. The danger ratings are further classified into three categories: no alert (less than 7)and high alert (greater than equal to 7). Our evaluation covers classical machine learning models, such as Support Vector Machines, as well as Neural Networks,  and transformer-based models. Model performance is assessed using standard metrics such as accuracy, F1-score, and mean absolute error (MAE), and the results are compared to identify the most robust approach. This research contributes to developing a more accurate and generalizable danger assessment framework for video-based risk detection.
\end{abstract}


\section{Introduction}
\label{sec:introduction}
In recent years, the rise of video content has significantly increased the demand for robust risk detection systems capable of accurately assessing dangerous situations in real-time \cite{5539872}. The ability to automatically predict the level of danger in video streams is essential for enhancing safety in security applications, such as surveillance, public safety, and autonomous systems\cite{gupta2024vidasvisionbaseddangerassessment}. Traditional methods \cite{10.1145/3584376.3584589} of assessing video content rely heavily on manual review, which is not scalable or efficient for large-scale deployments. As such, there is a growing need for automated systems that can quickly and accurately predict danger levels based on video data.

The field tends to emphasize "danger detection" methods rather than "danger assessment," which has limitations in terms of contextual generalization. Many existing approaches have a narrow focus, concentrating on specific detection techniques \cite{belsare2024context, wang2023improving, jin2020risk, de2019detection} or targeting particular types of dangers \cite{zhou2022student, wang2023improving, zhu2024deep, li2022pedestrian}, often ignoring the broader contextual elements essential for precise risk assessment. These non-generalized methods, especially in temporal visual scenarios, may overlook crucial factors like the broader danger context, relationships between objects, and localized actions over time—elements that are critical when evaluating the risk to humans in a scene. The widespread reliance on convolutional neural networks (CNNs) for danger and risk detection \cite{jean2021study, rajesh2020deep, marchella2023convolutional, wenqi2017model, Roy2024RoadAD, Iyavoo2024PerformanceAO} often necessitates extensive amounts of labeled data, which can be labor-intensive to gather but is vital for developing generalized danger assessment systems using CNNs.

In deep learning, understanding how a model perceives danger requires examining its approach to evaluating risk in a scene, highlighting that danger assessment can provide deeper insights than mere detection. Assessing danger in videos involves identifying hazardous elements and measuring the risk level they present, requiring more than simple object recognition or classification. It demands a comprehensive grasp of the context, interactions, and potential outcomes unfolding over time.

In this study, we make several key contributions to the field of video-based danger assessment:

1) Framework Development: We design and implement multiple machine learning and deep learning frameworks aimed at predicting danger levels in video content. These frameworks incorporate classical models like Support Vector Machines (SVMs) as well as modern deep learning architectures, including neural networks and transformer-based models.

2) Integration of Video and Text Embeddings: Our approach leverages CLIP \cite{radford2021learningtransferablevisualmodels} embeddings for video frames and GPT \cite{brown2020language} embeddings for textual summaries of the video content. We employ a strategy where embeddings from both modalities are combined to enhance the accuracy of danger prediction.

4) Comparative Model Analysis: We perform an extensive comparison of different models using a variety of embeddings. These comparisons include neural network-based binary classification models, SVM-based classifiers, and regression models using embeddings from CLIP and BERT.

Exploration of Model Architectures: Our study covers the design and evaluation of various architectures to determine their effectiveness in accurately predicting danger ratings in videos, contributing to more generalizable and scalable video-based risk detection methods.

\section{Related Works}

\textbf{Hazardous Scene Classification}: Hazardous scene classification involves the automatic identification of potentially dangerous situations in images or videos, such as accidents, riots, or assaults. Previous studies have primarily focused on creating datasets depicting hazardous scenarios \cite{5539872, mullen2024dontforgetmilkback, 9093457, 8578776} and developing methods for detecting these scenarios \cite{5539872, 10.1145/3584376.3584589}. The distinction between normal and dangerous scenes is often unclear \cite{8578776}, which can be addressed by evaluating danger on a standardized scale. This research is vital for enhancing public safety and preventing accidents.

\textbf{Temporal Action Localization (TAL)}: Temporal Action Localization aims to identify the temporal boundaries and spatial regions of specific actions in untrimmed videos. There are three primary approaches: one-stage pipelines that directly predict action boundaries \cite{8099638}, two-stage pipelines that first generate action proposals and then classify them \cite{Lee_Uh_Byun_2020}, and anchor-free methods for greater flexibility in action detection \cite{9633209}. Recent advances include Vision-Language Prompting, which uses natural language to enhance action localization \cite{9423269, nag2022zeroshottemporalactiondetection}.

Our work builds on TAL research by utilizing Vision-Language Prompting to interpret dangerous scenes, localize actions, and assess the danger level in videos.

\textbf{Zero-shot \& Few-shot Instruction}: Zero-shot and few-shot instruction learning enable LLMs to perform new tasks with little to no task-specific training, guided by prompt engineering \cite{brown2020language, coda2023meta, Li_2023, prabhumoye2022fewshotinstructionpromptspretrained}. These methods have proven effective in enhancing LLM performance, particularly in Moment Localization, which pinpoints specific moments in videos based on language queries \cite{gao2017talltemporalactivitylocalization, hendricks2017localizingmomentsvideonatural}. Our work extends this line of research by using zero-shot and few-shot instructions to guide LLMs in assessing danger levels in visual content.

\textbf{LLM-based Evaluators \& Meta-Evaluation}: LLM-based evaluators assess the performance of other LLMs, providing significant advantages in speed and scalability compared to traditional evaluation methods. Meta-evaluation involves assessing the reliability and validity of these evaluation methods. Although existing research demonstrates the potential of LLMs for automated evaluation \cite{lin2023llmevalunifiedmultidimensionalautomatic, liu2023calibratingllmbasedevaluator}, robust benchmarks for multimodal tasks like language-conditioned video analysis are still lacking. Our work addresses this gap by proposing a new benchmark for evaluating the danger assessment capabilities of vision-language models, contributing to the field through multimodal meta-evaluation aligned with human understanding.

\begin{figure}[t]
    \centering
    \includegraphics[width=\linewidth]{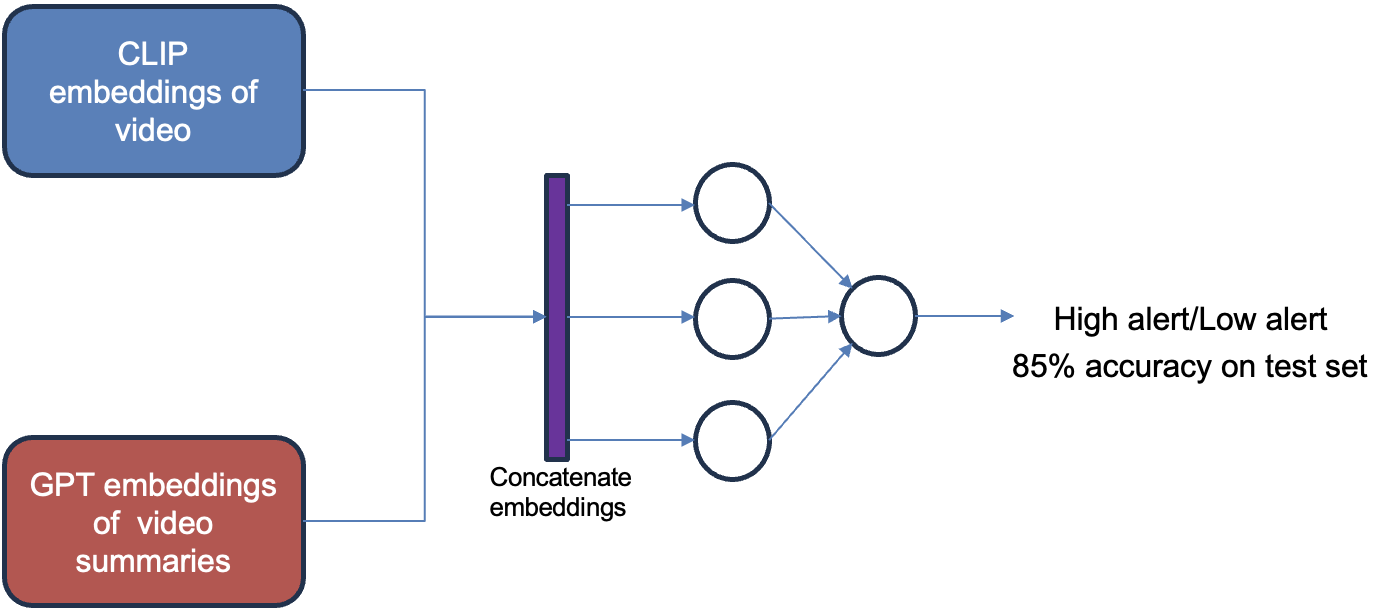}
    \caption{Concatenating CLIP and GPT Embeddings to classify a video if its high alert or not}
    \label{fig:arch1}
\end{figure}

\section{Methodology}

In this study, we compare various machine learning and deep learning models for predicting danger levels in videos, using ViDAS \cite{gupta2024vidasvisionbaseddangerassessment}, a dataset of 100 videos with human-annotated danger ratings ranging from 0 to 10. To effectively evaluate different approaches, we developed multiple frameworks, each combining visual and textual embeddings. Below, we detail all the frameworks used for this task.

\subsection{ViDAS Dataset}

The ViDAS dataset\cite{gupta2024vidasvisionbaseddangerassessment} consists of 100 videos. Each video is annotated with a danger rating between 0 and 10, determined by 18 human evaluators where each evaluator annotates each video. For classification purposes, we divide the ratings into two categories: \textit{high alert} (ratings $\geq$ 7) and \textit{no alert} (ratings $<$ 7).

For each of the experiments discusses in the paper we have selected 50 frames equally separated from a range determined by the temporal segment provided for each video in the dataset. These temporal segments are places where humans have agreed where the most danger lies in the video.

\subsection{CLIP and GPT Embeddings for Danger Classification}

This framework utilizes both visual and textual data to enhance prediction accuracy. The architecture of the model is shown in Figure \ref{fig:arch1}, and it consists of the following key steps:

\textbf{Video Embedding Extraction (CLIP)}: We extract visual embeddings from each video using the Contrastive Language-Image Pre-training (CLIP) model. CLIP captures both low- and high-level visual features from each frame, which are aggregated to form a single embedding representing the video’s visual content.
    
\textbf{Text Embedding Extraction (GPT)}: In parallel, we generate textual embeddings using a pre-trained GPT model. The text is derived from the summaries of the video in the dataset. GPT embeddings capture the semantic context of the scene and complement the visual features extracted by CLIP.
    
\textbf{Concatenation of Embeddings}: The CLIP embeddings of the video and GPT embeddings of the textual summary are concatenated to create a joint representation that includes both visual and semantic information.
    
\textbf{Classification Network}: The concatenated embeddings are passed through a fully connected neural network to classify the video into one of two categories: \textit{high alert} (ratings $\geq$ 7) or \textit{no alert} (ratings $<$ 7). The network is optimized using cross-entropy loss, and dropout is applied to prevent overfitting.
    
\textbf{Training and Testing}: The model is trained on 90\% of the dataset, with the remaining 10\% reserved for testing.

\begin{figure}[t]
    \centering
    \includegraphics[width=\linewidth]{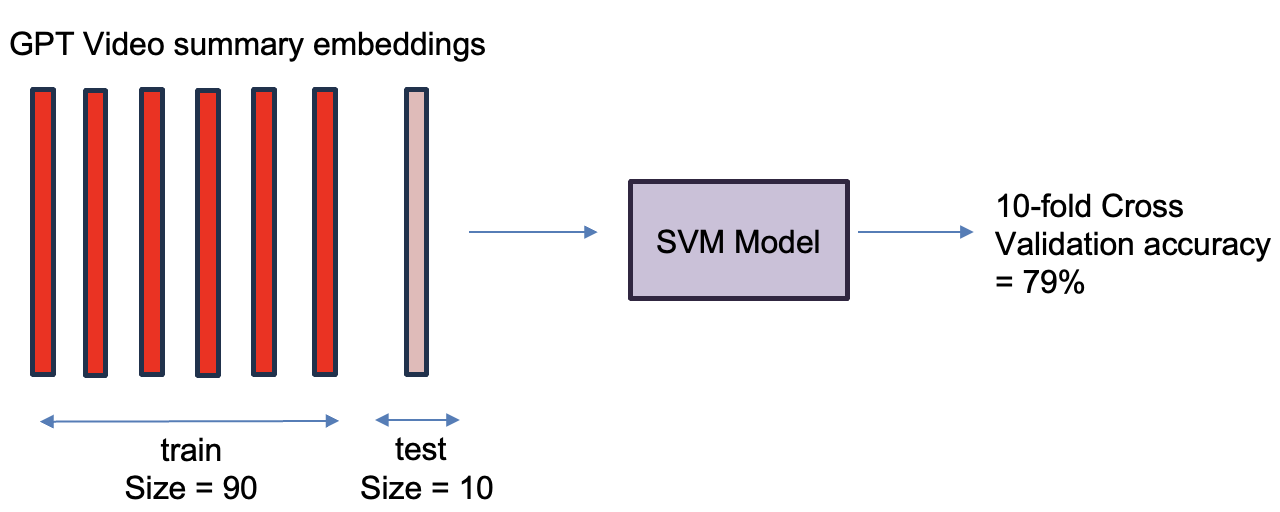}
    \caption{Perfoming classification using video summaries' embeddings using SVMs and 10-fold cross-validation strategy}
    \label{fig:arch2}
\end{figure}
\subsection{K-Fold SVM Cross-Validation with GPT Embeddings}

This framework focuses on using Support Vector Machines (SVMs) to classify danger levels in video content, based solely on GPT-generated textual embeddings of video summaries. The architecture of the framework is depicted in Figure \ref{fig:arch2}.

\textbf{Embedding Extraction (GPT)}: For this framework, we utilized GPT-generated embeddings from textual summaries of the videos. These summaries provide semantic context about the scenes and the associated danger levels. The embeddings serve as the primary input features for the SVM classifier.
    
\textbf{Train-Test Split}: The dataset of 100 videos is split into a training set of 90 videos and a test set of 10 videos. The embeddings are passed into the SVM model for training and classification.
    
\textbf{K-Fold Cross-Validation}: To evaluate the robustness of the model, we apply 10-fold cross-validation, where the dataset is divided into 10 subsets or "folds." The SVM model is trained on 9 folds and tested on the remaining fold. This process is repeated 10 times, ensuring that each fold is used for testing once. The final performance is averaged across all folds.

\begin{figure}[t]
    \centering
    \includegraphics[width=\linewidth]{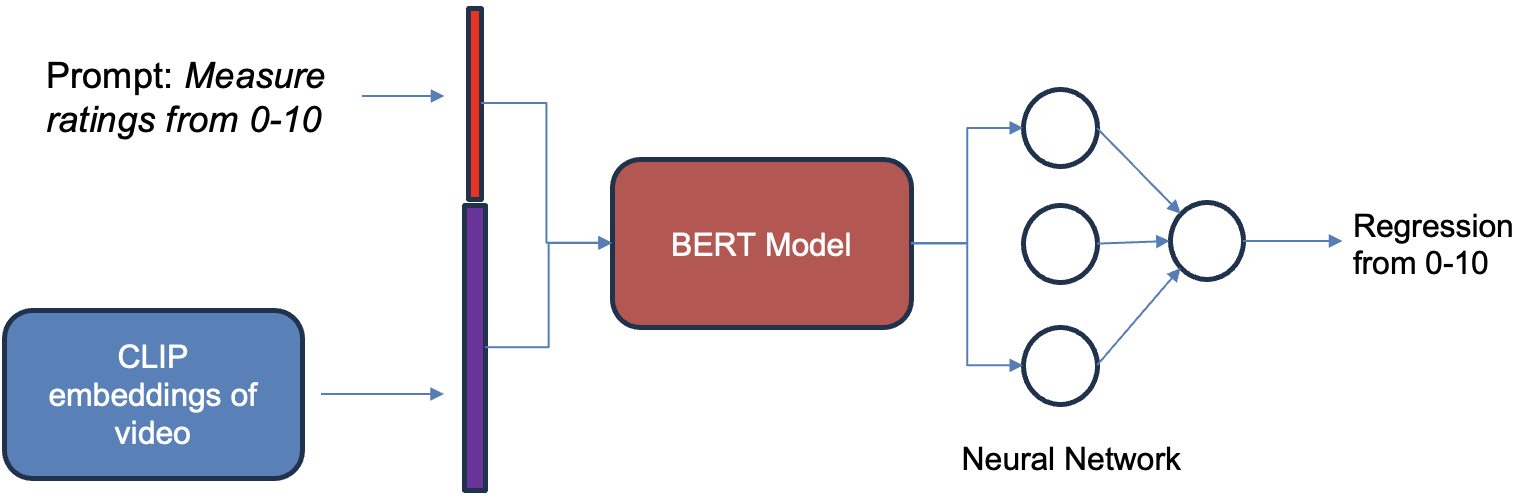}
    \caption{Concatenating CLIP and GPT Embeddings to classify a video if its high alert or not}
    \label{fig:arch3}
\end{figure}

\subsection{CLIP and BERT Embeddings for Danger Level Regression}

This framework introduces a regression-based approach to predict continuous danger ratings from 0 to 10. Unlike the previous classification-based frameworks, this model seeks to estimate the exact danger rating for each video using a combination of CLIP and BERT embeddings. The architecture of this framework is depicted in Figure \ref{fig:arch3}.

\textbf{Video Embedding Extraction (CLIP)}: Similar to the first framework, we use the CLIP model to extract visual embeddings from the video frames. These embeddings capture the visual characteristics and patterns related to danger levels in the video content.
    
\textbf{Text Embedding Extraction (BERT)}: To complement the visual information, we utilize the BERT model to generate embeddings from textual descriptions of the videos. BERT, a transformer-based language model, captures the contextual and semantic meaning of the provided text, enhancing the model's understanding of the video content.
    
\textbf{Concatenation of Embeddings}: The CLIP embeddings from the video and the BERT embeddings from the text are concatenated to form a unified feature vector. This feature vector serves as the input to the neural network responsible for predicting the danger rating.
    
\textbf{Neural Network Regression}: The concatenated embeddings are passed into a neural network, which performs regression to predict a continuous danger score ranging from 0 to 10. The neural network consists of fully connected layers and is trained using mean squared error (MSE) loss, a standard loss function for regression tasks.
    
\textbf{Training and Testing}: The model is trained on 90\% of the dataset, with the remaining 10\% reserved for testing. The performance of the regression model is evaluated using MSE , which quantify how well the predicted danger ratings align with the ground-truth human annotations.

\section{Results and Discussion}

In this section, we present the results obtained from each of the three frameworks described in the Methodology section. We analyze the performance of both the classification and regression models, using appropriate metrics such as accuracy, cross-validation accuracy, mean squared error (MSE). A discussion on the comparative performance of these models follows.

\subsection{CLIP and GPT Embeddings for Binary Classification}

\begin{figure} \label{fig:model1}
    \centering
    \includegraphics[width=0.8\linewidth]{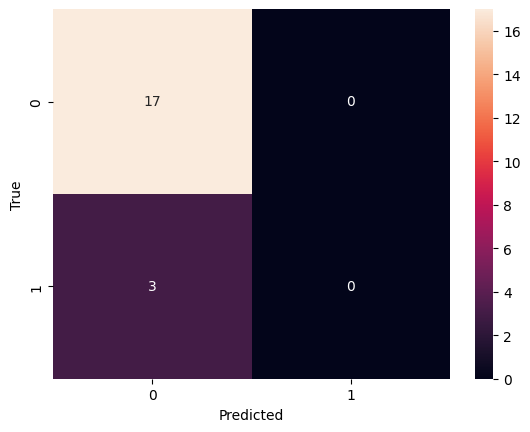}
    \caption{Confusion matrix depicting the performance of this framework.}
    \label{fig:enter-label}
\end{figure}

The first framework, which combines CLIP embeddings for video frames and GPT embeddings for textual summaries, achieved an \textbf{accuracy of 85\%} on the test set for the binary classification task (high alert or no alert). 

This result demonstrates the strength of combining visual and textual features for danger detection. The CLIP embeddings provide comprehensive visual context, while the GPT embeddings enrich the model’s understanding of the content, leading to more accurate classification.

\subsection{Framework 2: K-Fold SVM Cross-Validation with GPT Embeddings}

In the second framework, we utilized SVMs on GPT-generated embeddings and performed 10-fold cross-validation. The model achieved an average cross-validation accuracy of \textbf{79\%}, with individual fold accuracies ranging from 60\% to 90\%.

The high variance in fold accuracies indicates that the model's performance is dependent on the particular subset of data it is trained on, which could suggest potential biases or overfitting on certain aspects of the dataset. This framework demonstrates that textual embeddings alone can provide reasonable classification accuracy but may benefit from further tuning or the inclusion of visual data.

\subsection{CLIP and BERT Embeddings for Danger Level Regression}

The third framework focused on predicting continuous danger levels using regression, leveraging both CLIP (visual) and BERT (textual) embeddings. The model was evaluated using mean squared error (MSE).

\begin{itemize}
    \item \textbf{Mean Squared Error (MSE)}: 0.43

    \item \textbf{Discussion}: The regression framework allows for a more nuanced assessment of danger by providing a continuous score rather than a binary classification. The relatively low MSE indicate that the model was able to predict danger scores with a high degree of precision, although some variance still exists in certain edge cases. The combination of CLIP and BERT embeddings proved effective in capturing both visual and semantic aspects of the video, which is critical for precise danger rating predictions.
\end{itemize}

\subsection{Comparative Discussion}

Each framework offered distinct advantages and challenges:
\begin{itemize}
    \item The \textbf{first framework}, combining CLIP and GPT embeddings, achieved the highest classification accuracy (85\%) and demonstrated the power of fusing visual and textual data for danger detection.
    \item The \textbf{second framework} using SVMs and GPT embeddings provided reasonable performance (79\% cross-validation accuracy) but suffered from high variance across folds, suggesting that textual features alone may not be sufficient for reliable classification.
    \item The \textbf{third framework} for regression demonstrated strong predictive capabilities with low MSE values, showing promise for applications where continuous danger assessment is required.
\end{itemize}

Overall, the results suggest that models integrating both visual and textual embeddings outperform those relying solely on one modality. Moreover, the regression framework enables a more granular understanding of danger levels compared to the binary classification models.

\section{Limitations and Future Work}

While our work provides promising insights into video-based danger assessment, there are several limitations that need to be addressed to improve the generalizability and robustness of the proposed frameworks. First, the dataset used in this study is relatively small, consisting of only 100 videos with 50 frames each. This limited dataset size may not fully capture the variability and complexity of real-world danger scenarios, leading to potential overfitting in the models. Future work could involve the collection and annotation of a larger and more diverse dataset that encompasses a wider range of danger levels and contexts.

Another limitation lies in the reliance on pretrained models for feature extraction, such as CLIP and GPT. These models are not fine-tuned on our specific dataset, which may affect their ability to capture nuances in danger assessment tasks. Future research should explore the potential benefits of fine-tuning these models or developing domain-specific embeddings tailored for danger detection.

Additionally, the frameworks currently do not leverage temporal information between video frames effectively. Future work could involve incorporating recurrent neural networks (RNNs) or transformer-based temporal models to capture temporal dependencies, thereby improving the accuracy of danger assessments.

\section{Conclusion}
In this study, we presented a comparative analysis of three different machine learning and deep learning frameworks for predicting danger levels in video content. The first framework combined CLIP embeddings for video frames and GPT embeddings for textual summaries, achieving the highest accuracy of 85\% for binary classification. This highlighted the effectiveness of integrating visual and textual data for danger detection. The second framework employed SVMs on GPT embeddings alone, which achieved a cross-validation accuracy of 79\%. While reasonable, this approach demonstrated the limitations of relying solely on textual features without visual context. The third framework adopted a regression-based approach using CLIP and BERT embeddings to predict continuous danger scores, providing a nuanced assessment of danger levels with a low mean squared error.

Overall, our findings indicate that models combining visual and textual information outperform those that rely on a single modality. Additionally, the regression framework allowed for a more granular assessment of danger levels, which could be advantageous for real-world applications requiring continuous monitoring and assessment. Despite the limitations, the insights gained from this study lay the groundwork for future research in video-based risk detection, emphasizing the need for larger datasets, fine-tuned models, and the integration of temporal analysis to further enhance performance and generalizability.

\bibliographystyle{IEEEtran}  
\bibliography{paper}   
\vspace{12pt}

\end{document}